\documentclass{article}
\usepackage{ijcai23}

\usepackage{times}
\usepackage{soul}
\usepackage{url}
\usepackage[hidelinks]{hyperref}
\usepackage[utf8]{inputenc}
\usepackage[small]{caption}
\usepackage{graphicx}
\usepackage{amsmath}
\usepackage{amsthm}
\usepackage{booktabs}
\usepackage{algorithm}
\usepackage{algorithmic}
\usepackage{amssymb}
\usepackage{bbm}
\usepackage{multirow}
\usepackage{threeparttable}
\usepackage[switch]{lineno}

\title{Learning Survival Distribution with Implicit Survival Function}

\author{
Yu Ling
\and
Weimin Tan$^{*}$\And
Bo Yan$^{*}$ 
\affiliations
School of Computer Science, Shanghai Key Laboratory of Intelligent Information Processing, Shanghai Collaborative Innovation Center of Intelligent Visual Computing, Fudan University, Shanghai, China.
\emails
yling21@m.fudan.edu.cn,
\{wmtan, byan\}@fudan.edu.cn
\thanks{$^{*}$Corresponding author: Weimin Tan and Bo Yan. This work is supported by NSFC (Grant No.: U2001209, 61902076) and Natural Science Foundation of Shanghai (21ZR1406600). Our code is available at \url{https://github.com/Bcai0797/ISF}.}
}
\renewcommand\footnotemark{}

\begin{document}

\maketitle

\begin{abstract}
    Survival analysis aims at modeling the relationship between covariates and event occurrence with some untracked (censored) samples. In implementation, existing methods model the survival distribution with strong assumptions or in a discrete time space for likelihood estimation with censorship, which leads to weak generalization. In this paper, we propose Implicit Survival Function (ISF) based on Implicit Neural Representation for survival distribution estimation without strong assumptions, and employ numerical integration to approximate the cumulative distribution function for prediction and optimization. Experimental results show that ISF outperforms the state-of-the-art methods in three public datasets and has robustness to the hyperparameter controlling estimation precision.
\end{abstract}

\section{Introduction}

Survival analysis is a typical statistical task for tracking occurrence of the event of interest through modeling relationship between covariates and event occurrence. In some medical situations~\cite{DLpatientOutcome,DSCN4hi}, researchers model the death probability of some diseases using survival analysis to explore effects of prognostic factors. However, some samples lose tracking (censored) during observation. For example, some patients are still alive at the end of observation, whose survival times are unavailable. Such censored samples are valuable for analysis of favorable prognosis. Therefore, censorship is one key problem in survival analysis as well as survival distribution modeling.

The most widely-used survival analysis model Cox proportional hazard method~\cite{Cox1992} predicts a hazard rate, which assumes that the relationship between covariates and hazard is time-invariant. For optimization, Cox model and its extensions~\cite{lassoCox,MTLSA,DeepSurv,DeepConvSurv} maximize the ranking accuracy of comparable pairs including comparison between uncensored samples and censored samples.

Lately, some works introduce deep neural networks to survival analysis. DeepSurv~\cite{DeepSurv} and DeepConvSurv~\cite{DeepConvSurv} simply replace the linear regression in the Cox model with neural networks for non-linear representations. These methods maintain the strong assumption of hazards' time-invariance in Cox model, leading to weak generalization of networks in real-world applications. 

\begin{figure}[!t]
    \begin{minipage}[t]{1.0\linewidth}
      \centering
      \centerline{\includegraphics[height=3.5cm]{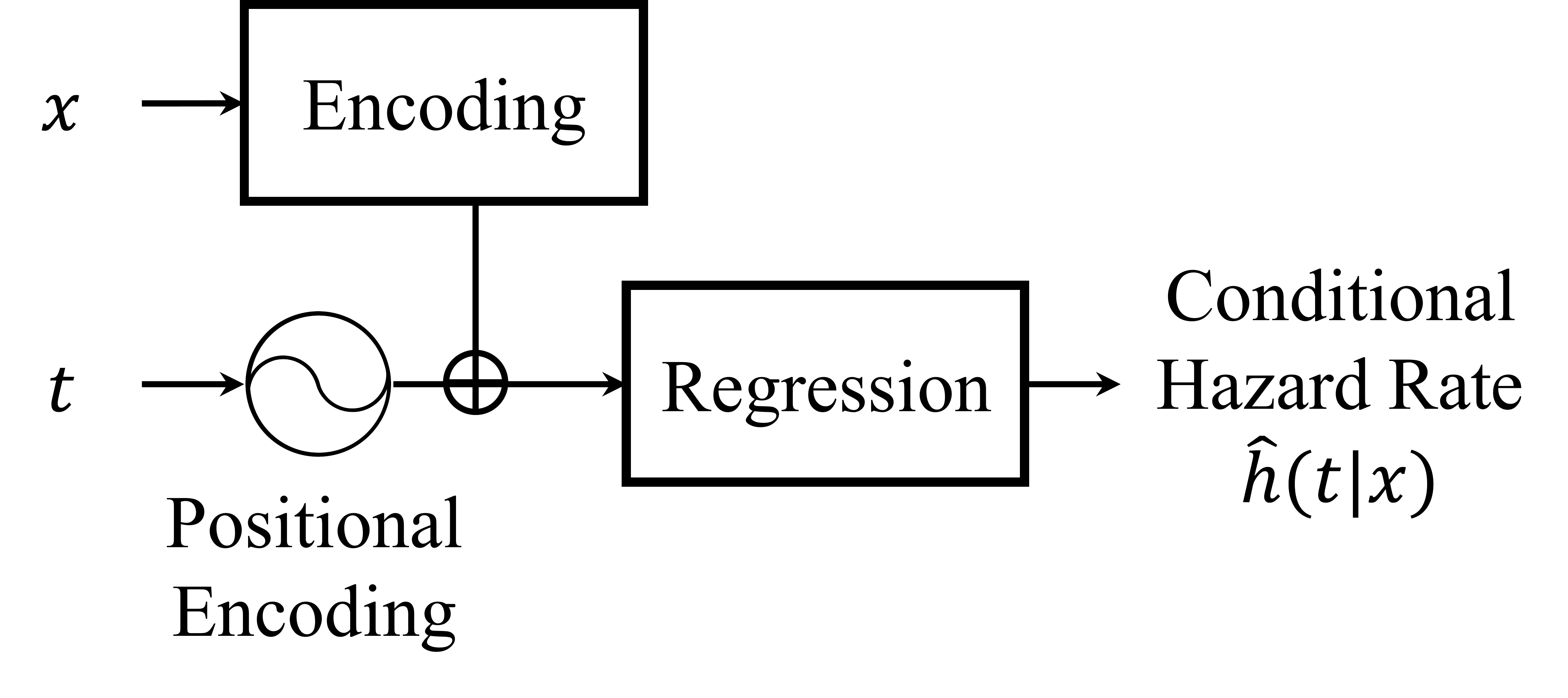}}
      \medskip
      \centerline{(a)}
      \medskip
    \end{minipage}
    \medskip
    \vfill
    \begin{minipage}[t]{1.0\linewidth}
      \centering
      \centerline{\includegraphics[height=4.12cm]{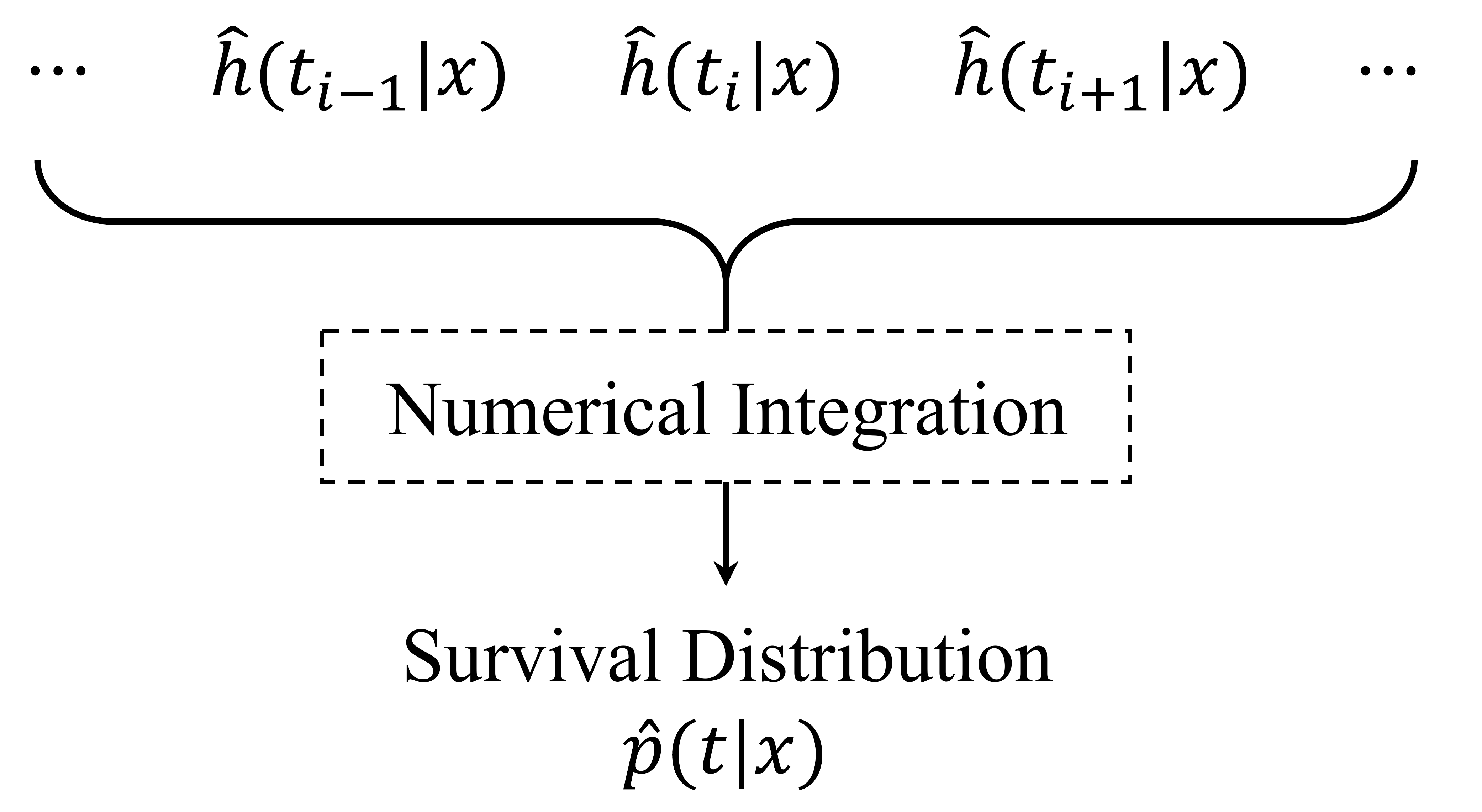}}
     \medskip
      \centerline{(b)}
    \end{minipage}  
    \caption{Brief framework of ISF. (a) ISF takes sample $x$ and time $t$ as input, and predicts conditional hazard rate $\hat{h}(t|x)$. (b) Based on estimated conditional hazard rates, we can derive survival distribution $\hat{p}(t|x)$ through numerical integration.}
    \label{fig:isf_brief}
\end{figure}

To avoid strong assumption on survival distribution, researchers try to estimate a distribution in a discrete time space instead of predicting a time-invariant risk. DeepHit~\cite{DeepHit} is proposed to learn occurrence probabilities at preset time points directly without assumptions about underlying stochastic process. Deep Recurrent Survival Analysis (DRSA)~\cite{DRSA} builds a recurrent network to capture the sequential patterns of the feature over time in survival analysis. Therefore, both DeepHit and DRSA learn a discrete survival distribution. Compared to the cross-entropy loss, the log-likelihood loss obtains better prediction for DeepHit and DRSA~\cite{biasCE}. On the basis of predicted occurrence probabilities in the discrete time space, the log-likelihood is naturally estimated in DeepHit and DRSA for both censored and uncensored samples.

Differing from discrete distribution estimation in DeepHit and DRSA, DSM~\cite{DSM} estimates the average mixture of parametric distributions. In implementation, DSM employs Weibull and Log-Normal distributions for analytical solutions of the cumulative distribution functions (CDF) and support limited in the space of positive reals. Therefore, DSM includes censored samples during optimization through CDF estimation. However, DSM also introduces assumptions on survival distribution through parametric distribution selection.

In this paper, we propose Implicit Survival Function (ISF) based on Implicit Neural Representation which is widely-used in 2D and 3D image representation~\cite{NeRFRS,LearningCIR}. As shown in Figure~\ref{fig:isf_brief}(a), ISF estimates a conditional hazard rate with the given sample and time. To capture time patterns, we embed the input time through Positional Encoding~\cite{Transformer}. The aggregated vector of encoded sample feature and time embedding is fed to a the regression module for conditional hazard rate estimation without strong assumptions on survival distribution. As shown in Figure~\ref{fig:isf_brief}(b), we employ numerical integration with predicted conditional hazard rates for survival distribution prediction. 

For optimization, we maximize likelihood of both censored and uncensored samples on the basis of approximated CDF of survival in a discrete time space. And experimental results prove that ISF is robust to the hyperparameter setting of the discrete time space.

To summarize, the contributions of this paper can be listed as:

\begin{itemize}
    \item The proposed Implicit Survival Function (ISF) directly models the conditional hazard rate without strong assumptions on survival distribution, and captures the effect of time through Positional Encoding.
    \item To estimate survival distribution with ISF, numerical integration is used to approximate the cumulative distribution function (CDF). Therefore, ISF can handle censorship common in survival analysis through maximum likelihood estimation based on approximated CDF.
    \item Though survival distribution estimation of ISF is based on a discrete time space, ISF has capability to represent a continuous survival distribution through Implicit Neural Representation. And experimental results show that ISF is robust to the setting of the discrete time space.
    \item To demonstrate performance of the proposed model compared with the state-of-the-art methods, experiments are built on several real-world datasets. Experimental results show that ISF outperforms the state-of-the-art methods.
\end{itemize}

\section{Formulation}
Survival analysis models aim at modeling the probabilistic density function (PDF) of tracked event defined as:
\begin{equation}\label{eq:p}
    p(t|x) = Pr(t_x=t | x)
\end{equation}
where $t$ denotes time, and $t_x$ denotes the true survival time.

Thus, the survival rate that the tracked event occurs after time $t_i$ is defined as:
\begin{align}\label{eq:S}
    S(t_i|x) & = Pr(t_x > t_i | x) \nonumber\\
    & = \int^{\infty}_{t_i} p(t|x)dt 
\end{align}

Similarly, the event rate function of time $t_i$ is defined as  the cumulative distribution function (CDF):  
\begin{align}\label{eq:W}
    W(t_i|x) & = Pr(t_x \leq t_i | x) = 1 - S(t_i | x) \nonumber \\
    & = \int^{t_i}_{0} p(t|x)dt
\end{align}

The conditional hazard rate $h(t|x)$ is defined as:
\begin{equation}\label{eq:h}
    h(t | x) = \lim_{\Delta t \rightarrow 0} \frac{Pr(t < t_x \leq t+\Delta t | t_x \geq t, x)}{\Delta t}
\end{equation}

\vspace{0pt}

\section{Related Work}\label{sec:related_work}

In this section, we describe several related approaches. The previous methods are divided into three parts based on their target of estimation: proportional hazard rate, discrete survival distribution and distribution mixture.  

\subsection{Proportional Hazard Rate}

The Cox proportional hazard method proposed in \cite{Cox1992} is a widely-used method in survival analysis tasks. Cox model assumes that the hazard rate of occurrence of a certain event is constant with time and the log of hazard rate can be represented by a linear function. Thus, the basic form of Cox model is:
\begin{align}\label{eq:h_cox}
\hat{h}(t | x) = h_0(t)exp(w^Tx)
\end{align}
where $t$ denotes time, $t_x$ denotes the true survival time, $x=(x_1, x_2, \dots, x_p)^T$ denotes covariates of samples, $w=(w_1, w_2, \dots, w_p)^T$  denotes parameters of the linear regression, and $h_0(t)$ denotes a fixed time-dependent baseline hazard function. Parameters $w$ can be estimated by minimizing the negative log partial likelihood.

However, the time-invariance assumption of hazard in Cox model weakens its generalization. Other methods make different assumptions about the survival function such as Exponential distribution~\cite{2003Statistical}, Weibull distribution~\cite{2016DSA}, Wiener process~\cite{Wiener} and Markov Chain~\cite{Markov}. These methods with strong assumptions about the underlying stochastic processes fix the form of survival functions, which suffers from generalization problem in real-world situations.

The outstanding capability of deep learning in non-linear regression achieve researchers' high attention. Therefore, many approaches introduce deep learning to survival analysis. DeepSurv~\cite{DeepSurv} replaces the linear regression of Cox model with a deep neural network for non-linear representation, but maintains the basic assumption of Cox model. Some works~\cite{DeepConvSurv,DCNN4sa} extend DeepSurv with a deep convolutional neural network for unstructured data such as images. 

\subsection{Discrete Probability Distribution}

To avoid strong assumptions about the survival time distribution, previous methods model the survival analysis problem in a discrete space with $K$ time points $T = \{ t^p_0, t^p_1, \cdots t^p_{k-1} \}$. DeepHit~\cite{DeepHit} uses a fully-connected network to directly predict occurrence probability $\hat{p}(t^p_i | x)$ defined as:
\begin{equation}\label{eq:p_deephit}
    \hat{p}(t^p_i | x) = Pr(t_x = t^p_i | x)
\end{equation}
where $t^p_i$ is a time point in the discrete time space $t^p_i \in T$.

DRSA~\cite{DRSA} employs standard LSTM units~\cite{LSTM} to capture sequential patterns of features over time, and predicts a conditional hazard rate defined as:
\begin{equation}\label{eq:h_drsa}
    \hat{h}(t^p_i | x) = \lim_{\Delta t \rightarrow 0} \frac{Pr(t^p_{i-1} < t_x \leq t^p_{i} | t_x \geq t^p_{i-1}, x)}{\Delta t}
\end{equation}

Hence, DRSA defines occurrence probability of event as:
\begin{equation}\label{eq:p_drsa}
    \hat{p}(t^p_i | x) = \hat{h}(t^p_i | x) \prod_{j<i} (1-\hat{h}(t^p_j | x))
\end{equation}

Although both DeepHit and DRSA predicts directly predict survival distribution without strong assumption, they only estimate probabilities at discrete time points.

\subsection{Distribution Mixture}

Discrete probability distribution estimation methods only estimate a fixed number of probabilities, which limits their applications. To generate a continuous probability distribution, DSM~\cite{DSM} learns a mixture of $K$ well-defined parametric distributions. Assuming that all survival times follows $t \geq 0$, DSM selects distributions which only have support in the space of positive reals. And for gradient based optimization, CDF of selected distributions require analytical solutions. In implementation, DSM employs Weibull and Log-Normal distributions, namely primitive distributions. 

During inference, parameters of $K$ primitive distributions $\left \{ \beta_k, \eta_k \right \}_{k=1}^K$ and their weights $\left \{ \alpha_k \right \}_{k=1}^K $ are estimated through MLP. Thus, the final individual survival distribution $\hat{p}(t | x)$ is defined as the weighted average of $K$ primitive distributions:
\begin{equation}\label{eq:dsm_p}
    \hat{p}(t | x) = \sum_{k=1}^K{\alpha_k P^p_k(t | x, \beta_k, \eta_k)}
\end{equation}

However, DSM introduces assumptions of survival distributions since primitive distribution selection is taken as a hyperparameter.

\section{Methodology}

\begin{figure}[!t]
    \begin{center}
       \includegraphics[width=0.8\linewidth]{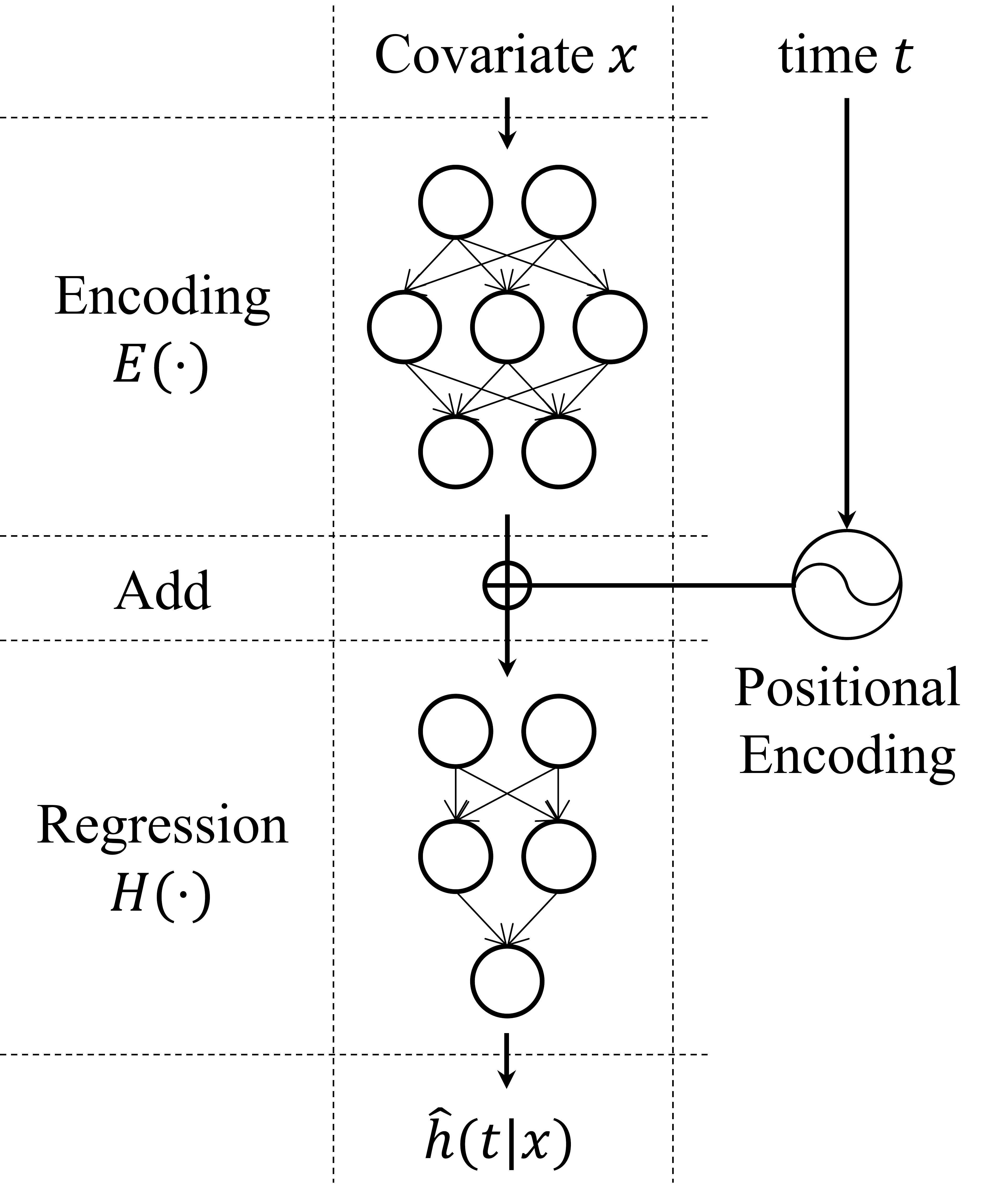}
    \end{center}
       \caption{Pipeline of ISF. Time $t$ is embedded through Positional Encoding ($PE$). Conditional hazard rate $\hat{h}(t|x)$ is estimated through $H(E(x)+PE(t))$, where $E(\cdot)$ and $H(\cdot)$ are implemented with MLP.}
    \label{fig:isf}
    \end{figure}

To model the survival distribution, we propose Implicit Survival Function (ISF) to estimate conditional hazard rate with positional encoding of time. In this section, we will demonstrate details of ISF as illustrated in Figure~\ref{fig:isf}.

\subsection{Implicit Survival Function}

The proposed ISF aims at predicting $h(t|x)$ defined in Eq.~\ref{eq:h}. For a given sample $x$, ISF first generates a feature vector $z \in \mathbb{R}^d$ using a Multilayer Perceptron (MLP) denoted by encoder $E(\cdot)$:
\begin{equation}\label{eq:z}
    z_x=E(x)
\end{equation}

To capture the effect of time, Positional Encoding ($PE$) of time $t$ is added to the feature vector $z$. Then, our hazard rate regression $\hat{h}(t|x)$ is defined as:
\begin{align}\label{eq:h_hat}
    \hat{h}(t|x) & = H(z_x + PE(t)) \nonumber \\
    & = H(E(x)+PE(t))
\end{align}
where $H(\cdot)$ is implemented with a MLP.

Positional Encoding maps time $t$ to a embedding of $d$ dimensions using pre-defined sinusoidal functions~\cite{Transformer}:
\begin{equation}\label{eq:PE}
    \left\{
    \begin{aligned}
        PE(t, 2i) &=sin(t/10000^{2i/d}) \\
        PE(t, 2i+1) &=cos(t/10000^{2i/d}) 
    \end{aligned}
    \right.
\end{equation}

The sinusoidal function based Positional Encoding provides shift-invariant representations, and let MLP learn high frequency functions~\cite{FourierFeature}. Therefore, ISF employs Positional Encoding defined in Eq.\ref{eq:PE} for embedding of time in survival analysis.

\subsection{Survival Distribution Estimation} \label{sec:surv_estimate}

For survival distribution estimation with ISF, we first estimate survival rate $S(t|x)$ defined in Eq.~\ref{eq:S}, and then approximate occurrence probability $p(t|x)$ defined in Eq.~\ref{eq:p} through difference of survival rate.

From Eqs.~\ref{eq:S} and \ref{eq:h}, we can derive the log survival rate at time $t_i$ as:
\begin{align}
    \ln S(t_i|x) & = \ln Pr(t_x > t_i | x) \nonumber \\
    & = \int_{0}^{t_i} \ln Pr(t_x > t | t_x \geq t, x) dt \nonumber \\
    & = \int_{0}^{t_i} \ln \left ( 1-h(t|x) \right ) dt 
\end{align}

Therefore, the estimated survival rate $\hat{S}(t_i | x)$ is defined as:
\begin{align}\label{eq:S_hat}
    \hat{S}(t_i | x) & = \exp \int_{0}^{t_i} \ln \left ( 1-\hat{h}(t|x) \right ) dt \nonumber  \\
    & = \exp \int_{0}^{t_i} \ln \left ( 1-H \left (E(x) + PE(t) \right ) \right ) dt
\end{align}

The estimated occurrence probability $\hat{p}(t | x)$ is approximated through:
\begin{align}\label{eq:p_hat}
    \hat{p}(t | x) & \approx Pr(t < t_x \leq t+\epsilon | x) \nonumber \\
     & \approx \hat{S}(t | x) - \hat{S}(t+\epsilon | x) 
\end{align}
where $\epsilon$ is a hyperparameter. The setting of $\epsilon$ depends on the precision of annotations in the dataset. Corresponding discussion is included in Section~\ref{sec:ablation_study}

For numerical stability, we manually set $\hat{S}(0|x)=1$ and $\hat{S}(t_{max} | x)=0$, where $t_{max}$ is ensured to be larger than any possible survival time in the dataset.

\subsection{Numerical Integration} \label{sec:integration}

Analytical solutions for integration in Eq.~\ref{eq:S_hat} is unavailable for ISF. To overcome such problem, we use numerical integration to approximate CDF in a discrete time space.

The duration of survival time $[0, t_{max})$ is split into $K$ intervals $\{ (t_i^p, t_{i+1}^p]\}^{K-1}_{i=0}$ with time points $T=\{t_i^p\}^{K}_{i=0}$, where $t_0^p = 0$ and $t_k^p = t_{max}$. In this paper, we set $t_{i+1}^p = t_i^p + \epsilon$ for convenience.

Let $g(t, x)$ denote $\ln(1-\hat{h}(t|x))$. Therefore, the integration in Eq.~\ref{eq:S_hat} for $t_i^p \in T$ is calculated using Simpson Formula as:
\begin{align}\label{eq:simpson}
    \hat{S}(t_i^p | x) & = \exp \int_{0}^{t_i^p} g(t, x) dt \nonumber  \\
    & \approx \exp \sum_{j < i} \frac{\epsilon}{6} [ g(t_j^p, x) + 4 g(t_j^p + \frac{\epsilon}{2}, x) + g(t_{j+1}^p, x) ]
\end{align}

Thus, the event rate (CDF) is estimated as $\hat{W}(t_i^p | x) = 1 - \hat{S}(t_i^p | x)$.

\subsection{Loss Function} \label{sec:loss}

Like existing approaches~\cite{DeepHit,DRSA,DSM}, we construct loss functions on the basis of maximum likelihood estimation. Although ISF provides a conditional hazard rate in the continuous time space, the optimization is performed in the discrete time space for CDF approximation. In this section, for easily understanding, we describe the proposed loss function separately for censored and uncensored samples in the view of predicting $\hat{p}(t|x)$, though forms of loss functions for these two types of samples are the same. 

\subsubsection{Censored Samples}

For a censored sample, the true survival time $t_x$ is unknown but the latest observation time $t_x^o$ is available, which indicates $t_x > t_x^o$. Thus, the loss function is expected to maximize $\hat{S}(t_x^o | x)$. For simplification, we maximize $\hat{S}(t_i^p | x)$ where $t_x^o \in (t_i^p, t_{i+1}^p]$. 

Therefore, the loss function for censored samples is defined as:
\begin{align}\label{eq:L_cs}
    L_ {cs}(x) & = - \ln \hat{S}(t_i^p | x) \nonumber \\
    & = - \ln \sum_{j \geq i} \hat{p}( t_j^p | x) 
\end{align}
where the latest observation time $t_x^o \in (t_i^p, t_{i+1}^p]$.

\subsubsection{Uncensored Samples}

Given an uncensored sample $(x, t_x^o)$, the observation time $t_x^o$ is equal to the true survival time $t_x$. Thus, we maximize $\hat{p}(t_i^p | x)$ where the true survival time $t_x^o \in (t_{i-1}^p, t_i^p]$:
\begin{equation}\label{eq:L_ucs}
    L_ {ucs}(x) = - \ln \hat{p}(t_i^p | x)
\end{equation}

\subsubsection{Unified Loss}

According to $L_{cs}$ in Eq.~\ref{eq:L_cs} and $L_{ucs}$ in Eq.~\ref{eq:L_ucs}, loss for both uncensored and censored samples can be represented as sum of $\hat{p}(t_i^p|x)$ in the discrete time space. For unification, we first define an indicator vector $Y^x \in \mathbb{R}^K$ in the discrete time space including $K+1$ time points as:
\begin{equation}\label{eq:Y}
Y^x_i = 
\left \{ 
\begin{matrix}
1 & t_x^o \in (t_i^p, t_{i+1}^p] \\
0 & t_x^o \notin (t_i^p, t_{i+1}^p]
\end{matrix}
\right.
\end{equation}

Thus, the proposed loss function can be unified as:
\begin{equation}\label{eq:L}
L(x) = - \ln \left( Y^x_i  \hat{p} \left( t_i | x \right) \right)
\end{equation}

The unified loss function $L(\cdot)$ handles both censored and uncensored samples. We use indicator vector $Y^x$ to control likelihood calculation. Hence, the proposed loss function is suitable for any type of censorship. 

\subsection{Computational Complexity} \label{sec:compute_complex}

As discussed in Sections~\ref{sec:surv_estimate}, \ref{sec:integration} and \ref{sec:loss}, estimation and optimization of ISF is performed in a discrete time space with $K$ time intervals. For $N$ samples, ISF predicts $O(NK)$ occurrence probabilities for survival distribution estimation. However, such process can be accelerated in the parallel computation situation because of independent positional encoding of time points.

\subsection{Difference from Existing Methods}

\begin{figure*}[!t]
    \begin{minipage}[t]{.48\linewidth}
      \centering
      \centerline{\includegraphics[width=5.10cm]{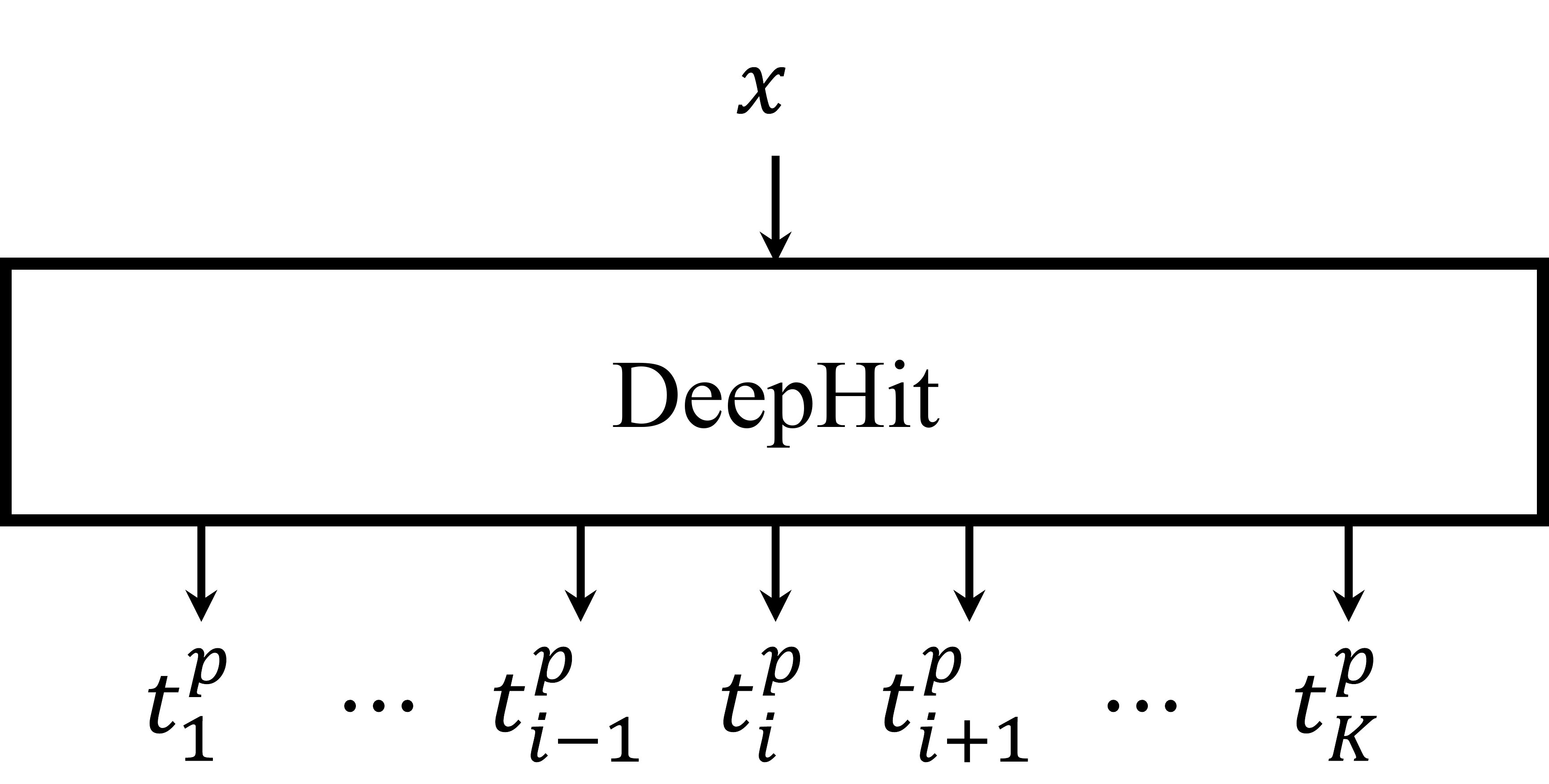}}
      \medskip
      \centerline{(a)}
    \end{minipage}
    \begin{minipage}[t]{.48\linewidth}
      \centering
      \centerline{\includegraphics[height=2.08cm]{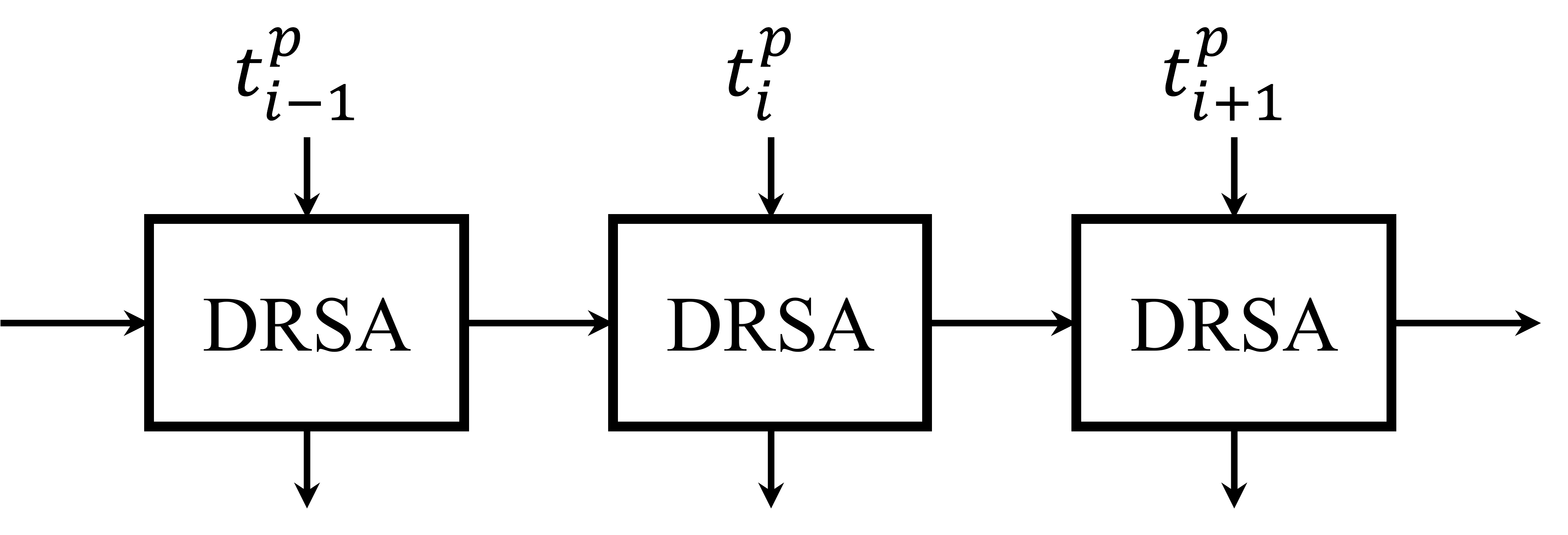}}
     \medskip
      \centerline{(b)}
    \end{minipage}  
    \vfill
    \begin{minipage}[t]{.48\linewidth}
      \centering
      \centerline{\includegraphics[width=5.10cm]{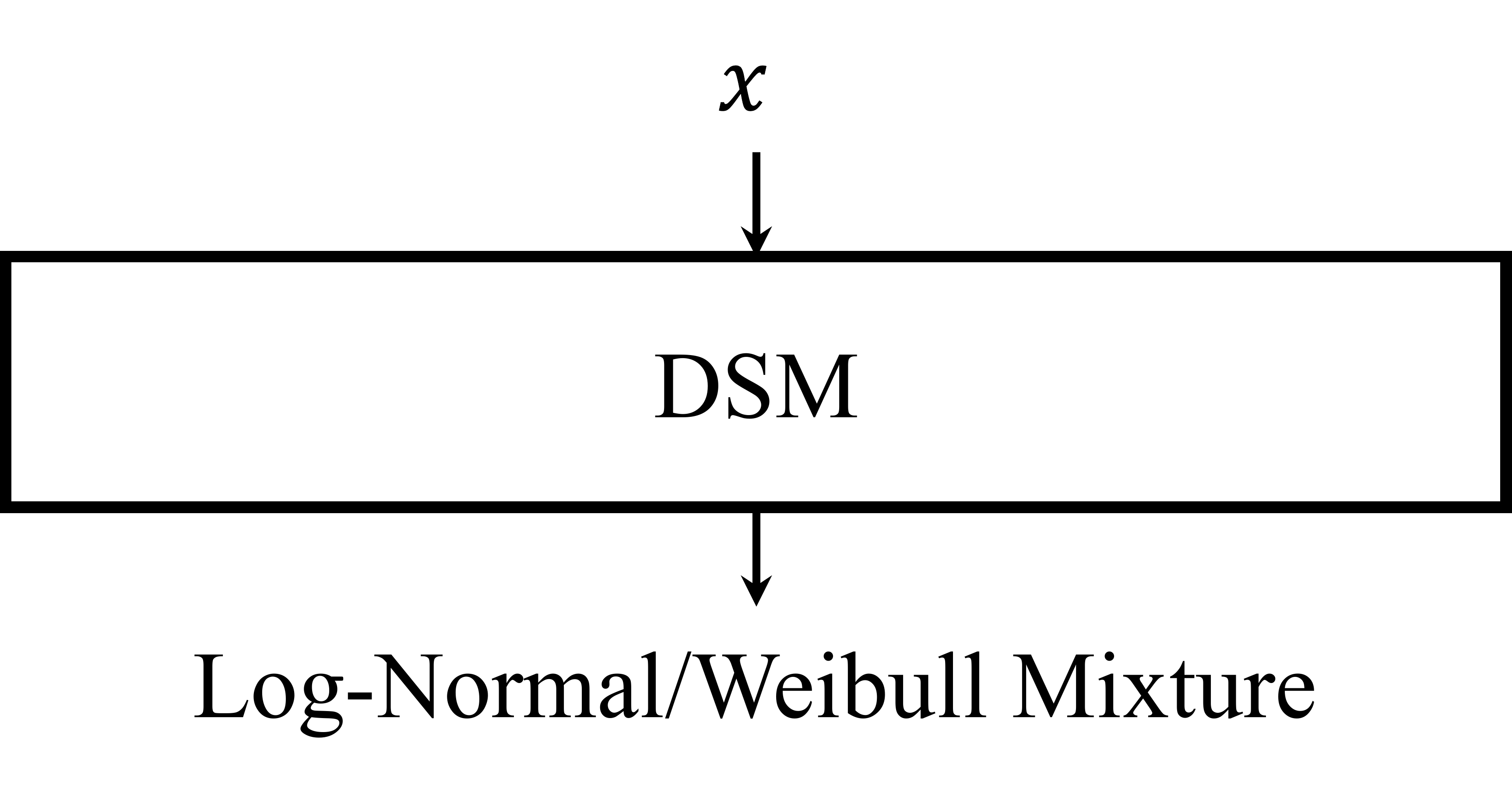}}
      \medskip
      \centerline{(c)}
    \end{minipage}
    \begin{minipage}[t]{.48\linewidth}
      \centering
      \centerline{\includegraphics[height=2.08cm]{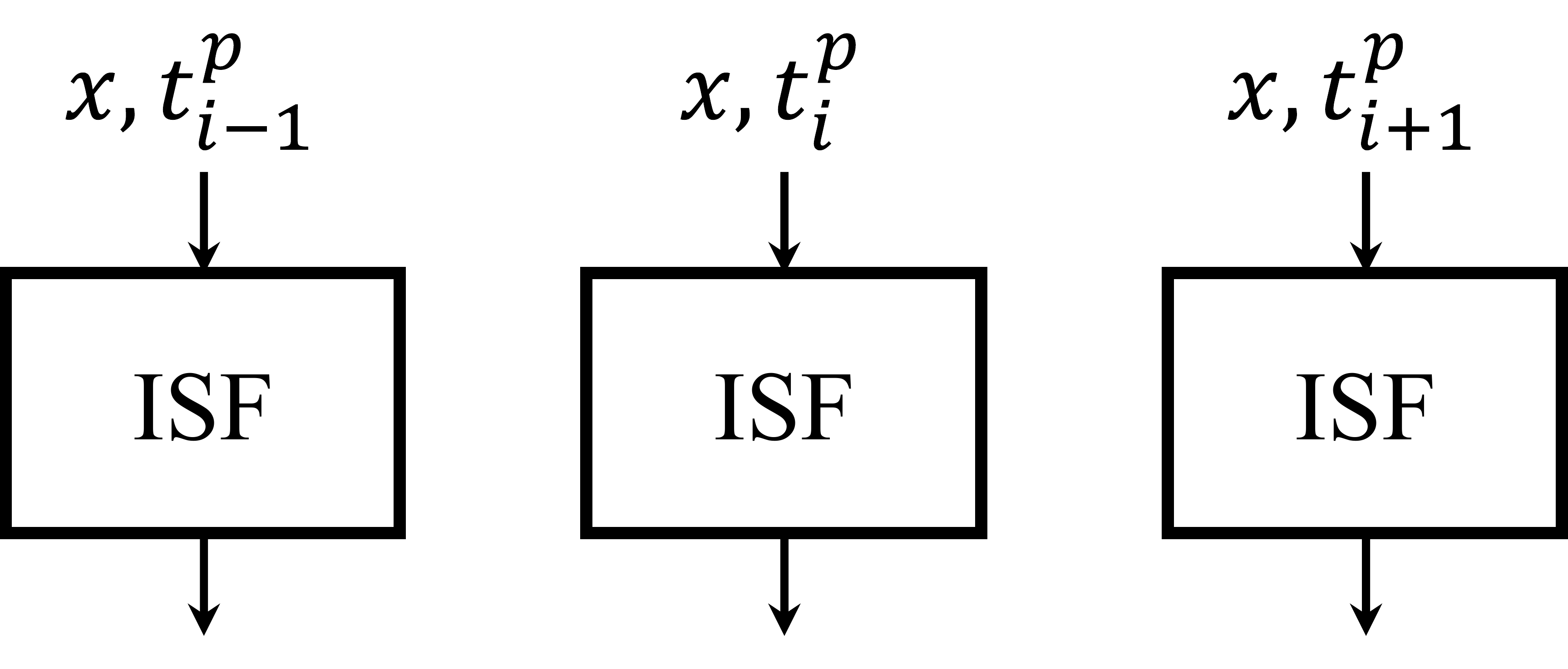}}
      \medskip
      \centerline{(d)}
    \end{minipage}
    \caption{Framework comparison between existing methods and ISF. (a) DeepHit predicts occurrence probabilities at preset time points. (b) RNN based DRSA sequentially estimates conditional hazard rates over time. (c) DSM models survival distribution through estimates parameters of mixture of parametric distributions (Log-Normal/Weibull). (d) ISF takes sample $x$ and time $t^p_i$ as input, and generates independent estimation for time points.}
    \label{fig:model_diff}
\end{figure*}

In this section, we compare the proposed model ISF with deep-learning models DeepHit, DRSA and DSM whose survival distribution estimation is close to that of ISF. We illustrate brief frameworks of these models and ISF in Figure~\ref{fig:model_diff}.

\subsubsection{ISF vs DeepHit}

As shown in Figure~\ref{fig:model_diff}(a), DeepHit directly regresses occurrence probabilities at preset time points through MLP. Therefore, the number of parameters dependents on the number of time points in the discrete time space. 

Since ISF takes positional encoding of time as input, the number of parameters in ISF is independent to the amount of time points. Therefore, ISF has better expansibility for time space variation.

\subsubsection{ISF vs DRSA}

According to Eqs.~\ref{eq:h_drsa} and \ref{eq:h_hat}, the goal of both ISF and DRSA is conditional hazard rate estimation. With estimated hazard rate, occurrence probability can be easily derived as shown in Eqs.~\ref{eq:p_drsa} and \ref{eq:p_hat}. 

The main difference between ISF and DRSA is the method of capturing time effect. As shown in Figure~\ref{fig:model_diff}(b), DRSA applies RNN to learn sequential patterns in a discrete time space and serially processes preset time points, while ISF uses positional encoding to exploit time information in the real field through parallel computation.

\subsubsection{ISF vs DSM}

DSM models continuous survival distribution with mixture of parametric distributions as shown in Figure~\ref{fig:model_diff}(c). Instead of explicit distribution representation in Eq.~\ref{eq:dsm_p}, ISF learns a function $H(\cdot)$ taking time as input defined in Eq.~\ref{eq:h_hat} to directly estimate conditional hazard rate. Therefore, the implicit representation of survival distribution in ISF avoids strong assumptions on survival distribution. 

With decrease of $\epsilon$ in Eq.~\ref{eq:p_hat}, precision of occurrence probability approximation increase, and thus ISF can be regarded as approximation of a continuous survival distribution. Distribution mixture in DSM directly models a continuous survival distribution, but distribution selection is a hyperparameter with strong assumptions about the stochastic process.

\section{Experiments}

\begin{table*}[tb]
    \centering
    \begin{tabular}{l|rrrrr}
        \toprule
        Dataset & \#Total Data & \#Censored Data & Censoring Rate & \#Features & Max Time \\
        \midrule
        CLINIC & 6,036 & 797  & 0.132 & 14 & 82\\
        MUSIC & 3,296,328 & 1,157,572  & 0.351 & 6 & 300 \\
        METABRIC & 1,981 & 1,093  & 0.552 & 21 & 356 \\
        \bottomrule
    \end{tabular}
    \caption{The statistics of CLINIC, MUSIC and METABRIC.}
    \label{tab:public_dataset}
\end{table*}

In this section, we compare the proposed method ISF with the state-of-the-art deep-learning survival distribution estimation methods including DeepHit, DRSA and DSM. DeepHit predicts the occurrence probability $\hat{p}(t | x)$ directly with a fully-connected neural network~\cite{DeepHit}. DRSA estimates a conditional hazard rate $\hat{h}(t | x)$ with LSTM units to capture sequential patterns~\cite{DRSA}. Both DeepHit and DRSA perform survival analysis in the discrete time space, while DSM estimates a continuous survival distribution through the mixture of parametric distributions~\cite{DSM}. Besides, we also compare ISF with Cox~\cite{Cox1992}, its deep-learning extension DeepSurv~\cite{DeepSurv} and random forest based survival analysis method RSF~\cite{RSF}.

\subsection{Datasets}

  To demonstrate the performance of the proposed method, experiments are conducted on several public real-world dataset:
  
  \begin{itemize}
     \item \textbf{CLINIC} tracks patients' clinic status~\cite{support}. The tracked event is the biological death. Survival analysis in CLINIC is to estimate death probability with physiologic variables.
     \item \textbf{MUSIC} is a user lifetime analysis containing about $1000$ users with entire listening history~\cite{MUSIC}. The tracked event is the user visit to the music service. The goal of survival analysis is to predict the time elapsed from the last visit of one user to the next visit.
     \item \textbf{METABRIC} dataset contains gene expression profiles and clinical features of the breast cancer from 1,981 patients~\cite{METABRIC}. Following the experimental setting of DeepHit, 21 clinical features are used during evaluation~\cite{DeepHit}. 
  \end{itemize}
  
  The statistics of three datasets is shown in Table~\ref{tab:public_dataset}. The training and testing split of CLINIC and MUSIC follows the setting of DRSA~\cite{DRSA}. For METABRIC, 5-fold cross validation is applied following DeepHit~\cite{DeepHit}.

\subsection{Metric}

Concordance Index (C-index, CI) is a widely-used evaluation metric in survival analysis for measuring the probability of accurate pair-wise order of comparable samples' event time. However, the ordinary CI~\cite{CI} for proportional hazard models assumes the predicted value is time-invariant~\cite{Cox1992,lassoCox,DeepSurv}, while distribution estimation based methods predict a time-dependent distribution of survival. Thus, following DeepHit and DSM, we perform time-dependent concordance index~\cite{CItd}, which is defined as:
\begin{equation}\label{eq:CItd}
    CI = Pr \left( W(t_{x_i} | x_i) > W(t_{x_j} | x_j ) | t_{x_i} < t_{x_j} \right) 
\end{equation}
where $t_{x_i}$ denotes the true survival time of $x_i$.

\subsection{Implementation Details}

For fair comparison, the discrete time space in experiments is set as $\left \{ (0, 1], (1, 2], \dots, (K-1, K] \right \}$ following setting of DeepHit and DRSA. According to the maximum time shown in Table~\ref{tab:public_dataset}, $t_{max}$ is set as $400$, and $K=t_{max}$.

ISF is implemented with $PyTorch$. Number of hidden units of $E(\cdot)$ defined in Eq.~\ref{eq:z} and $H(\cdot)$ defined in Eq.~\ref{eq:h_hat} are corresponding set as $\{256, 512, 256\}$ and $\{256, 256, 1\}$ for all experiments.

During training, we perform Adam optimizer. Models of the best CI is selected with variation in hyperparameters of learning rate $\{10^{-3}, 10^{-4}, 10^{-5}\}$, weight of decay $\{10^{-3}, 10^{-4}, 10^{-5}\}$ and batch size $\{8, 16, 32, 64, 128, 256\}$. The influence of $\epsilon$ will be discussed in the ablation study.

The reproduction of DeepHit and DRSA is based on the official code of DRSA\footnote[1]{https://github.com/rk2900/drsa}. And the reproduction of DSM refers to the official package $auto\_survival$\footnote[2]{https://autonlab.github.io/auton-survival/models/dsm}.

\subsection{Performance Comparison}

\begin{table}[!htp]
    \centering
	\small
    \begin{tabular}{l|ccc}
        \toprule
        \multirow{2}*{Method} & \multicolumn{3}{c}{CI $\uparrow$} \\
        \cmidrule{2-4}
        ~ & CLINIC & MUSIC & METABRIC \\
        \midrule
        \multirow{2}*{Cox} & 0.525$^\ddagger$ & 0.524$^\ddagger$ & 0.648$^\ddagger$\\
		~ & (0.512-0.538) & (0.523-0.525) & (0.634-0.662)\\
        \multirow{2}*{RSF}  & 0.598$^\ddagger$ & 0.566$^\ddagger$ & 0.672$^\ddagger$\\
		~ & (0.594-0.602) & (0.565-0.567) & (0.655-0.689) \\
        \multirow{2}*{DeepSurv}  & 0.532$^\ddagger$ & 0.578$^\ddagger$ & 0.648$^\ddagger$\\
		~ & (0.519-0.545) & (0.574-0.582) & (0.636-0.660)\\
        \multirow{2}*{DeepHit}  & 0.586$^\ddagger$ & 0.550$^\ddagger$ & 0.677$^\ddagger$\\
		~ & (0.567-0.605) & (0.549-0.551) & (0.665-0.688)\\
        \multirow{2}*{DRSA} & 0.580$^\ddagger$ & 0.610$^\ddagger$ & 0.692$^\dagger$\\
		~ & (0.564-0.596) & (0.601-0.619) & (0.672-0.712)\\
        \multirow{2}*{DSM} & 0.598$^\ddagger$ & 0.593$^\ddagger$ & 0.697$^*$ \\
		~ & (0.582-0.613) & (0.579-0.606) & (0.677-0.718) \\
        \multirow{2}*{ISF} & \textbf{0.612} & \textbf{0.701} & \textbf{0.704} \\
		~ & (0.596-0.629) & (0.700-0.702) & (0.681-0.728)\\
        \bottomrule
        \multicolumn{4}{p{210pt}}{$*$: $p\geq0.05$, $\dagger$: $p<0.05$, $\ddagger$: $p<0.01$; unpaired t-test with respect to ISF.} 
    \end{tabular}
    \caption{Comparison of CI (mean and 95\% confidence interval) in four public datasets CLINIC, MUSIC and METABRIC.}
    \label{tab:performance}
\end{table}

To evaluate performance of ISF, we conduct experiments in three public datasets CLINIC, MUSIC and METABRIC compared with several existing methods. Since compared discrete time space methods DeepHit and DRSA set time points as $t^p_{i+1}=t^p_{i}+1$, $\epsilon$ in Eq.~\ref{eq:p_hat} which controls precision of ISF is set as $1$ during training and evaluation for fair comparison.

As shown in Table~\ref{tab:performance}, ISF achieve the best CI in three datasets which censoring rates are $0.132$, $0.351$ and $0.552$. Therefore, ISF is robust to censoring rate. Besides, the large number of samples in MUSIC dataset contributes to performance improvement of ISF, while ISF has relatively low improvement in METABRIC containing fewer samples. 

\subsection{Ablation Study} \label{sec:ablation_study}

For further understanding of ISF, we conduct experiments on ISF with variation of $\epsilon$ in Eq.~\ref{eq:p_hat} which controls precision to study the effect of precision. As discussed in Section~\ref{sec:compute_complex}, ISF predicts $O(NK)$ occurrence probabilities for $N$ samples with $K$ time intervals where $K \propto 1/\epsilon$. 

\subsubsection{Training Precision}

\begin{table}[!tb]
    \centering
    \begin{tabular}{ccc|c|ccc}
        \toprule
        \multicolumn{7}{c}{Training $\epsilon$} \\
        1/10 & 1/5 & 1/2 & 1 & 2 & 5 & 10 \\
        \midrule
        0.613 & 0.614 & 0.613 & 0.612 & 0.613 & 0.611 & 0.600 \\
        \bottomrule
    \end{tabular}
    \caption{CI performance comparison with variation of $\epsilon$ during training in CLINIC. During inference, $\epsilon$ of all models is fixed to $1$ for fair comparison and accurate evaluation.}
    \label{tab:ablation_train_epsilon}
\end{table}

Since survival time annotations in CLINIC are saved as integer, the ideal $\epsilon$ for CLINIC is $\epsilon=1$. Therefore, we evaluate CI of ISF on CLINIC with variation of $\epsilon$ during training in this section. For fair comparison and accurate evaluation, $\epsilon$ in inference in this section is fixed to $\epsilon^{Inference}=1$.

As defined in Eq.~\ref{eq:p_hat}, $\epsilon$ determines precision of ISF. In CLINIC dataset, estimation precision of ISF is higher than annotation precision when $\epsilon^{Train} < 1$ during training. On the contrary, if $\epsilon^{Train} > 1$, annotation precision is higher than estimation precision. In such case, ISF predicts occurrence probabilities at unseen time points.

In Table~\ref{tab:ablation_train_epsilon}, results of $\epsilon^{Train}$ from $0.1$ to $10$. For $\epsilon^{Train} \in [0.1, 1)$, ISF achieves close CI since estimation precision of these models is higher than annotation precision. For $\epsilon^{Train} \in \{2, 5\}$, the performance is also close to that of ISF with $\epsilon^{Train}=1$, which indicates that ISF is capable of extrapolating in a certain range of time and robust to $\epsilon^{Train}$ variation. In the extreme case of $\epsilon^{Train}=10$, CI of ISF significantly decreases since the maximum survival time in CLINIC is $82$. 

\subsubsection{Inference Precision}

\begin{table}[!tb]
    \centering
    \begin{tabular}{l|ccc|c}
        \toprule
        \multirow{2}*{Dataset} & \multicolumn{4}{c}{Inference $\epsilon$} \\
        ~ & 1/10 & 1/5 & 1/2 & 1 \\
        \midrule
        CLINIC & 0.609 & 0.610 & 0.612 & 0.612 \\
        MUSIC & 0.695 & 0.696 & 0.698 & 0.701 \\
        METABRIC & 0.703 & 0.703 & 0.704 & 0.704 \\
        \bottomrule
    \end{tabular}
    \caption{CI performance comparison with variation of $\epsilon$ during training in CLINIC, MUSIC and METABRIC. The evaluated ISF is trained with $\epsilon=1$.}
    \label{tab:ablation_eval_epsilon}
\end{table}

In this section, we study generalization ability of ISF with variation of $\epsilon^{Inference}$ during evaluation. Based on ISF trained with $\epsilon^{Train}=1$, we adjust $\epsilon^{Inference}$ from $0.1$ to $1$ during inference, and evaluate corresponding CI performance in three public datasets. In $\epsilon^{Inference}<1$ experiments, ISF predicts conditional hazard rates at time points unseen in training. Hence, results of CI demonstrate generalization ability of ISF. 

As shown in Table~\ref{tab:ablation_eval_epsilon}, ISF performance has little decrease when $\epsilon^{Inference}<\epsilon^{Train}$. Hence, ISF has high generalization for occurrence probability prediction at time points beyond the preset discrete time space, which proves that ISF manages to capture patterns of time through representations from sinusoidal positional encoding. 

\section{Discussion}

In this section, we discuss some features of ISF in details.

\subsection{Estimation Precision}

In this paper, We use a hyperparameter $\epsilon$ to control the sampling density of the discrete time space, which has impact on the estimation precision of ISF. Experimental results of the ablation study in Section~\ref{sec:ablation_study} show that ISF with varied $\epsilon$ achieves close CI performance in a certain range, even if the estimation precision is lower than annotation precision.

ISF captures time patterns through positional encoding as defined in Eq.~\ref{eq:PE}. Representation based on sinusoids is shift-variation and enables MLP learn high frequency functions~\cite{FourierFeature}. Therefore, ISF manages to extrapolate occurrence probabilities unseen during training. 

Although low $\epsilon$ leads to high computational complexity as discussed in Section~\ref{sec:compute_complex}, the generation ability of ISF enables models trained with relatively high $\epsilon$ to generate acceptable results of survival prediction.

\subsection{Discrete Time Space}

ISF estimates conditional hazard rates in a discrete uniform time space for optimization and inference. For $N$ samples with $K$ time intervals, ISF processes $O(NK)$ pairs of sample and time during training and inference. In this section, we discuss the necessity of uniform time sampling.

In Section~\ref{sec:loss}, we maximize occurrence probabilities at time points $t^p_i$ instead of observed time $t^o_x$. If ISF maximizes $\hat{p}(t^o_x|x)$ or $\hat{S}(t^o_x | x)$ during optimization, the number and distribution of processed sample-time pairs depends on the training set. In the extreme case that the training set contains $N$ samples with highly discrete survival time, ISF processes $O(N^2K)$ sample-time pairs with numerical integration in $K$ intervals for optimization based on $t^o_x$. And the distribution of these sample-time pairs relies on the distribution of observed time, which perhaps introduces prior of the survival time distribution in the training set. Though ISF based on the discrete time space replaces the observed time with preset time points, the optimization process is based on adjustable uniform sampling of time. And the adjustment of the discrete time space is independent to the model architecture of ISF.

The ablation study of $\epsilon$ also proves that the preset discrete uniform time space based optimization and inference provides enough accuracy for survival analysis. Moreover, the estimation precision of ISF can be easily changed without model architecture modification through variation of hyperparameter $\epsilon$. Hence, occurrence probabilities prediction in a discrete time space through ISF like previous works~\cite{DeepHit,DRSA} is reasonable and robust.

\subsection{Unified Loss Function}

In real-world applications, right-censoring is most common in datasets, which indicates that the true survival time is larger than the observed time $t^x > t^o_x$. Therefore, existing discrete or continuous distribution prediction methods only considers right-censoring in loss functions~\cite{DeepHit,DRSA,DSM}. 

Instead of establishing two distinct loss functions for censored and uncensored samples, the proposed loss function uses indicator vector $Y$ defined in Eq.~\ref{eq:Y} for likelihood calculation. Therefore, a unified loss function defined in Eq.~\ref{eq:L} is proposed for both censored and uncensored samples and is easy to be extended for any type of censoring.

\section{Conclusion}

In this paper, we propose Implicit Survival Function (ISF) for conditional hazard rate estimation in survival analysis. ISF employs sinusoidal positional encoding to capture time patterns. Two MLP are used to encode input covariates and regress conditional hazard rates. For survival distribution estimation, ISF performs numerical integration to approximate CDF for survival rate prediction.

Compared with existing methods, ISF estimates survival distribution without strong assumptions about survival distribution and models a continuous distribution through Implicit Neural Representation. Therefore, ISF models based on different settings of the discrete time space share a common architecture of the network. Moreover, ISF has robustness to estimation precision controlled by the discrete time space whether the estimation precision is higher than the annotation precision or not. Experimental results show that ISF outperforms the state-of-the-art survival analysis models on Concordance Index performance in three public datasets with varied censoring rates.

\newpage

\bibliographystyle{named}
\bibliography{ijcai23}

\end{document}